\documentclass{article}

\usepackage[preprint]{nips_2018}

\usepackage[utf8]{inputenc} % allow utf-8 input
\usepackage[T1]{fontenc}    % use 8-bit T1 fonts
\usepackage{hyperref}       % hyperlinks
\usepackage{url}            % simple URL typesetting
\usepackage{booktabs}       % professional-quality tables
\usepackage{amsfonts}       % blackboard math symbols
\usepackage{amssymb}       
\usepackage{nicefrac}       % compact symbols for 1/2, etc.
\usepackage{microtype}      % microtypography
\usepackage{amsthm}
\usepackage{amsmath}
\usepackage{natbib}
\usepackage{graphicx}
\usepackage{subcaption}
\newtheorem{lemma}{Lemma}
\newtheorem{theorem}{Theorem}
\DeclareMathOperator{\E}{\mathop{{}\mathbb{E}}}
\title{A Fourier View of REINFORCE}

\author{
  Adeel Pervez \\
  Faculty of Computer Science\\
  GIK Institute, Pakistan\\
  \texttt{adeel@giki.edu.pk} \\
}
\setcitestyle{numbers}
\setcitestyle{square}

\begin{document}

\maketitle

\begin{abstract}
  We show a connection between the Fourier spectrum of Boolean functions and the REINFORCE gradient estimator for binary latent variable models. We show that REINFORCE estimates (up to a factor) the degree-1 Fourier coefficients of a Boolean function. Using this connection we offer a new perspective on variance reduction in gradient estimation for latent variable models: namely, that variance reduction involves eliminating or reducing Fourier coefficients that do not have degree 1.  We then use this connection to develop low-variance unbiased gradient estimators for binary latent variable models such as sigmoid belief networks. The estimator is based upon properties of the noise operator from Boolean Fourier theory and involves a sample-dependent baseline added to the REINFORCE estimator in a way that keeps the estimator unbiased. The baseline can be plugged into existing gradient estimators for further variance reduction.
\end{abstract}

\section{Introduction}
Gradient-based optimization is the workhorse of contemporary machine learning. For deterministic models, the backpropagation algorithm has allowed computing of gradients relative to parameters of large neural network models. Recent work has sought to extend this success to models where some of the hidden variables are stochastic. The \emph{reparameterization trick} \cite{vae} has enabled efficient training of models that have continuous stochastic hidden variables by allowing computation of unbiased and low-variance gradients of loss functions. Models with discrete stochastic variables, although attractive from a modeling point of view in terms of their power and interpretability, have proven difficult to optimize using gradient-based optimization. Neither backpropagation nor the reparametrization trick are directly applicable to training discrete variable models. Optimizing discrete variables models using REINFORCE and REINFORCE-like algorithms is difficult due to high-variance of these gradient estimators.

In this paper we use ideas from the theory of harmonic analysis of Boolean functions to provide another perspective on the REINFORCE \cite{williams} gradient estimator. Harmonic analysis seeks to analyze Boolean functions using \emph{Fourier expansions} which express Boolean functions as linear combinations of orthonormal basis functions with coefficients called the \emph{Fourier coefficients} of the Boolean function. Each Fourier coefficient corresponds to a subset of input coordinates and depends on the product probability distribution imposed on the Boolean cube. We show that REINFORCE estimates singleton Fourier coefficients of a Boolean function. We then introduce the noise operator from harmonic analysis and construct a control variate that can be plugged into existing gradient estimators to further reduce their variance.

The contribution of this paper is two-fold.
\begin{enumerate}
\item The first is a reinterpretation of existing variance-reducing gradient estimators in terms of the Fourier coefficients of the objective function seen as a function of the stochastic hidden variables. We show that they can be seen as eliminating or reducing terms with irrelevant Fourier coefficients from the Fourier expansion of the objective function.
\item The second contribution is an unbiased estimator for the gradient of the objective function relative to the parameters of the stochastic binary latent variables. The estimator depends on the properties of the noise operator from Boolean Fourier theory.
\end{enumerate}

We begin with a review of the Fourier analysis of Boolean functions.

\section{Background}
\subsection{Boolean Fourier Analysis}
We reiterate some facts from the analysis of Boolean functions. A comprehensive introduction can be found in \cite{odonnell}. We work with Boolean functions $f:\{-1,1\}^n \to \mathbb{R}$.  We assume a product probability distribution on the Boolean input, $p(x)=\prod_{i=1}^n p_i(x)$ with $p_i$ being the probability of the $i$th coordinate being one, $\mu_i$ its mean and $\sigma_i^2$ its variance. Given Boolean functions $f$ and $g$, we define an inner product: 
\[\langle f,g \rangle = \E_{p(x)}[f(x) g(x)].\] 

We define the $p$-norm of $f$:
\[ \|f\|_p = \E[|f(x)|^p]^{1/p}, \]
with the property that $\|f\|_p \le \|f\|_q $ if $p \le q$.

\subsubsection{Fourier Expansion}
Let $\phi_i(x)=\frac{x_i-\mu_i}{\sigma_i}$. For a set $S \subseteq [n]$, define $\phi_S(x) = \prod_{i\in S} \phi_i(x)$. The $2^n$ functions $\phi_S$ are an orthonormal basis for the space of Boolean functions. The expansion of $f(x)$ in this basis
\[ f(x) = \sum_{S \subseteq [n]} \hat{f}(S)\phi_S(x) \]
is known as the $p$-biased Fourier expansion of $f$ and the coefficients of this expansion, $\hat{f}(S)$, are the Fourier coefficients. The cardinality of $S$ is the \emph{degree} or \emph{weight} of the coefficient. The Fourier expansion of a Boolean function expresses the function as a multilinear polynomial in $\phi_i(x_i)$. It should be noted that the Fourier coefficients of a function depend on both that function and the product probability distribution imposed on the input and changing the input probability distribution changes the Fourier expansion for the same function.  

There is also an inverse expansion for the Fourier coefficients
\[ \hat{f}(S) = \langle f,\phi_S \rangle = \E[f(x) \phi_S(x)]. \]
%\rho correlated inputs
%noise operators
In particular, the degree-$0$ coefficient is the average of the function under the input distribution: $\hat{f}(\varnothing) = \E[f(x)]$. 

\subsubsection{Discrete Derivative}
The $i$th discrete derivative of $f$ is formal derivative of the $p$-biased expansion of $f$ relative to $\phi_i$. This is denoted by $D_i f(x)$ and has the Fourier expansion given by
\[ D_if = \sum_{S \ni i} \hat{f}(S)\phi_{S\setminus {i}}(x)\]
Notice that $D_if$ does not depend on $x_i$.

\subsubsection{Noise Operator}
Given $x,x' \in \{-1,1\}^n$ and $\rho \in [0,1]$, we say that $x, x'$ are $\rho$-correlated if $x'$ is generated by independently setting each $x'_i$ to $x_i$ with probability $\rho$ and a sample from $p_i$ with probability $1-\rho$. 
\[
    x'_i =  \begin{cases}
            x_i &\text{ with probability } \rho \\
            \text{ random sample from } p_i & \text{ with probability } 1-\rho \\
            \end{cases}
    \]

We also denote this by $x' \sim N_\rho(x)$. We use this to define the \emph{noise operator} $T_\rho$ acting on $f$ as the expectation over $\rho$-correlated inputs as follows:
\[ T_\rho(f)(x) = \E_{x' \sim N_\rho(x)}[f(x')] \]
The noise operator has a special action on the functions $\phi_i$: it multiplies them by $\rho$, i.e., 
\[T_\rho(\phi_i)(x) = \E_{x' \sim N_\rho(x)}[\phi_i(x')] =  \rho \phi_i(x). \]
Similarly, its action on the basis function $\phi_S$ is to multiply it by $\rho^{|S|}$:
\[T_\rho(\phi_S)(x) =T_\rho\left(\prod_{i\in S} \phi_i\right)(x)=\prod_{i\in S} T_\rho(\phi_i)(x)= \rho^{|S|} \phi_S(x). \]
By linearity of expectation it follows that
\[ T_\rho(f)(x) = \sum_{S \subseteq [n]} \hat{f}(S)T_\rho(\phi_S)(x) = \sum_{S \subseteq [n]} \rho^{|S|} \hat{f}(S)\phi_S(x). \]

\section{Related Work}
Perhaps the best known example of a gradient estimator for (but not limited to) discrete stochastic hidden variable models is the REINFORCE algorithm \cite{williams}. Also known as the likelihood ratio estimator and the score function estimator, this estimator uses the log-derivative trick i.e., $\frac{\partial}{\partial \theta_i} p_\theta(x) = p(x)\frac{\partial \log p(x)}{\partial \theta_i}$, to convert the gradient of an expectation with respect to parameters of a probability distribution over which the expectation is computed into an expectation of a term with a gradient of a log probability. That is,
\[ \frac{\partial}{\partial \theta_i} \E_{p(x|\theta)}[f(x)] = \E_{p(x)}\left[f(x) \frac{\partial}{\partial \theta_i} \log p(x|\theta)\right] \] 
The score function estimator has been applied to optimize the variational lower bound on the log-likelihood with an inference network \cite{vae, rezende}. In this form, however, the estimator has high variance which leads to slow convergence. A number of subsequent methods have been proposed to deal with this high variance.  For continuous stochastic variables \cite{vae} propose the reparameterization trick which involves rewriting the function in a form that allows gradients to go through.

Another method to reduce the variance of this estimator, which also works for discrete variables, is to subtract a term $c$, called a \emph{control variate}, from $f(x)$. If the control variate depends on the samples, $c=c(x)$, then we must add the analytical expectation of the control variate under $p(x)$ to keep estimator unbiased. 
\[ \frac{\partial}{\partial \theta_i} \E_{p(x|\theta)}[f(x)] = \E_x\left[(f(x)-c(x)) \frac{\partial}{\partial \theta_i} \log p(x|\theta)\right] + \E_x\left[c(x) \frac{\partial}{\partial \theta_i} \log p(x|\theta)\right] \] 
A number of control variate schemes have been proposed: NVIL \cite{nvil} subtracts two baselines from the objective to reduce variance: the first is a constant baseline set to the moving average of the function and the second is an input-dependent baseline computed by a feedforward neural network. Since the baselines do not depend on the samples, the analytical expectation is identically 0. MuProp \cite{muprop} uses the first-order Taylor approximation of the function $f(x') + f'(x')(x'-x)$ as a baseline. $x'$ is usually set to the distribution mean $\mu$ which allows gradients to pass through but requires a separate pass through the network. Since the baseline depends on $x$, the term $f'(\mu)\E[x]$ must be added to keep the estimator unbiased. DARN \cite{darn} also uses the first-order Taylor expansion as a baseline but does not add the analytical expectation, making the estimator biased. Another very simple biased estimator is the straight-through estimator \cite{st} which uses the gradient relative to the sample as that relative to the parameter. Another class of estimators employ multiple samples such as in \cite{burda} to construct tighter lower-bounds on the log likelihood and in \cite{vimco} to construct per-sample baselines.

The Fourier expansion is widely used in computational learning theory with applications to learning low-degree functions \cite{mansour}, decision trees \cite{decision}, constant-depth circuits \cite{lmn}, juntas \cite{junta}.

\section{REINFORCE Estimates Fourier Coefficients}
The following lemma is a slight variant of the Margulis-Russo \cite{margulis,russo, odonnell} formula. See the appendix for proof.
\begin{lemma}
    \label{lemma:1}
    Let $f$ be a Boolean function. Then
    \[ \frac{\partial}{\partial p_i} \E_{p_\theta (x)}[f(x)] = \frac{2}{\sigma_i}\hat{f}(\{i\}), \]
    where $\hat{f}(\{i\}) = \E_{p(x)}[f(x) \phi_{i}(x)]$ is the Fourier coefficient of $f$ under the $p$-biased expansion.
\end{lemma}

From this we see that:
\begin{align}
    g_{\textsubscript{REINFORCE}}^i &= \E_{p(x)}\left[f(x) \frac{\partial}{\partial p_i}\log p(x)\right] \\
                                &= \frac{2}{\sigma_i}\hat{f}(\{i\})\\
                                &= \E_{p(x)}\left[f(x) \frac{2\phi_{i}(x)}{\sigma_i}\right].
\end{align} 

\section{Variance Reduction in Gradient Estimation}
From the above we see that computing gradients of loss functions with respect to parameters of binary latent variables requires computing weight-1 Fourier coefficients. Since a Fourier coefficient is an expectation, Monte Carlo averaging is frequently used as an estimation technique. Naive averaging however results in an estimator that has high variance. What is the source of this variance? To see this we can use the Fourier expansion as follows. A Fourier coefficient is written
\begin{align}
    \hat{f}({i}) &= \E[f(x)\phi_i(x)] = \E\left[\left(\sum_{S\subseteq [n]} \hat{f}(S)\phi_S(x)\right)\phi_i(x)\right]\\
     &= \hat{f}(\varnothing)\E[\phi_i(x)] + \hat{f}({i})\E[\phi_i^2(x)] + \sum_{\substack{S\subseteq [n]\\S \notin \{\varnothing,\{i\}\}}} \hat{f}(S)\E\left[\phi_S(x)\phi_i(x)\right].
\end{align}
Since $\E[\phi_i(x)] = \E\left[\phi_S(x)\phi_i(x)\right] = 0$ and $\E[\phi_i^2(x)] = 1$ because of orthonormality, we end up with $\hat{f}({i})$ in expectation. But using a Monte Carlo average to estimate the expectation will get a contribution to its variance coming from the first and third terms. This variance will be high or low if the Fourier coefficients in these terms are high or low respectively. From this perspective, to reduce the variance of a gradient estimator we should attempt to remove those terms with irrelevant Fourier coefficients or ensure that those Fourier coefficients are small in magnitude.

In the following we look at some gradient estimators from the literature with this view.

\subsection{Straight-Through Estimator}
The straight-through estimator \cite{st} is a simple but biased estimator that computes that derivative with respect to a sample and uses it as an estimate of the derivative with respect to the probability parameter $p_i$.
\[ g_{\textsubscript{ST}}^i =  \frac{\partial f(x)}{\partial x_i} \]

We can view this estimator as approximating the discrete derivative of $f$ or equivalently the derivative of the Fourier expansion of $f$. The discrete derivative of $f$ can be expanded as follows.
\[ D_i(f)(x_{\setminus i}) = \sum_{S\ni i} \hat{f}(S) \phi_{S\setminus i}(x_{\setminus i}) \]
Here $x_{\setminus i}$ denotes $x$ without the $i$th coordinate. If we were able to compute $D_i$ we would have an unbiased estimator of the gradient since $\E[D_i(f)(x)] = \widehat{D_i(f)}(\varnothing) = \hat{f}(\{i\}).$ However, since we normally have $f$ available to us as a neural network (and not as a  multilinear polynomial), we approximate the discrete derivative with the derivative of the function as a neural network.

\subsection{NVIL}
NVIL \cite{nvil} subtracts two baselines, a constant baseline and an input dependent baseline from the function before computing its Fourier coefficient. Since the baselines do not depend on samples from the distribution, they can be seen as subtracting the weight-0 Fourier coefficient, $\hat{f}(\varnothing)$, from the function. Notice that this does not affect the remaining Fourier coefficients of the resulting function.

\subsection{MuProp}
The MuProp \cite{muprop} estimator builds a control variate based on the first order Taylor series expansion as follows.

\[ g_{\mu}^i = \left(f(x) - f(\mu) - \frac{\partial f(\mu)}{\partial \mu_i}(x_i-\mu_i)\right)\frac{\partial \log p(x)}{\partial p_i}+ \frac{\partial f(\mu)}{\partial \mu_i}\frac{\partial}{\partial p_i}\E[x_i] \]

This can be viewed as an approximation to the following.
\[ g_{\mu}^i = \left(f(x) - \hat{f}(\varnothing) - D_if(x_{\setminus i})\frac{x_i-\mu_i}{\sigma_i}\right)\frac{2\phi_i(x_i)}{\sigma_i}+ \frac{2D_if(x_{\setminus i})}{\sigma_i} \]

The term $D_if(x_{\setminus i})\frac{x_i-\mu_i}{\sigma_i}$ includes all the terms of $f$ that contain $x_i$ and subtracting from $f$ removes all terms containing $x_i$ from $f$, reducing the variance in $f$. Once again, since we do not have $f$ as a multilinear polynomial, we cannot compute $D_if(x)$, therefore we approximate it with the gradient of the function as a neural network.

\section{Proposed Estimator}
We develop an unbiased gradient estimator using the variance reduction properties of the noise operator. The noise operator, $T_\rho$, is a smoothing operator that when applied to a functions decays the effect of the higher-order terms of the function. The extent of decay also depends on degree: the higher the degree of a term, the greater is the decaying factor applied to the term. The expected value of a function under the input distribution is unaffected by application of the noise operator i.e., $\E[f] = \E[T_\rho(f)]$. Therefore the noise operated version of a function $f$ is a function $T_\rho(f)$ with the same expected value but with much reduced variance.

This is intuitively clear from the Fourier expansion of the noise operator with parameter $\rho$:
\[ T_\rho(f)(x) = \sum_{S \subseteq [n]} \rho^{|S|} \hat{f}(S)\phi_S(x) \]
Here, we see that the noise operator multiplies the  Fourier coefficients of $f$ with an exponentially decaying factor in $\rho^{|S|}$ that depends on the degree of the term. We also see that as $\rho \rightarrow 0$, $T_\rho(f) \rightarrow \E[f]$ and as $\rho \rightarrow 1$, $T_\rho(f) \rightarrow f$. In other words, the variance of $T_\rho(f)$ goes to $0$ with $\rho$. For intermediate values of $\rho$, $T_\rho(f)(x)$ has exponentially small higher degree terms and lower variance than $f$. In fact, we can make the stronger statement that even higher norms of $T_\rho$ are bounded by the second norm of $f$. This fact is expressed by saying that the noise operator is \emph{hypercontractive} and is the content of Bonami's hypercontractivity \cite{bonami, odonnell} theorem. One version of the theorem states

\begin{theorem}[Hypercontractivity of Noise Operator]
    Given a product distribution on a finite probability space where each outcome in the constituent distributions has probability at least $\lambda$ and a function $f$ defined on such space, for any $q>2$ and $0\le\rho\le\frac{1}{\sqrt{q-1}}\lambda^{1/2 - 1/q}$, we have that 
    \[\|T_\rho f\|_q \le \|f\|_2. \]
\end{theorem}

In particular for appropriate $\lambda$ and $\rho$ we have that $ \|T_\rho f\|_2\le \|T_\rho f\|_{2+\epsilon} \le \|f\|_2 $. Here we use that fact that for $p \le q$, $\|f\|_p \le \|f\|_q$. Squaring and subtracting the square of the expectation $\E[f]^2 = \E[T_\rho(f)]^2$, we see that 
\[ \|T_\rho f\|^2_2-\E[f]^2\le \|f\|^2_2-\E[f]^2\]
\[ \sigma^2(T_\rho(f)) \le \sigma^2(f), \]
showing that the variance of the noise operated version of $f$ is no more than the variance of $f$.

\subsection{Constructing the Control Variate}
We can use this fact to construct a control variate in a number of ways. Recall that a control variate for an unbiased estimator is essentially a Boolean function for which all degree-1 coefficients are 0. If $g(x)$ is any Boolean function then 
\[g(x) -\frac{1}{\rho}T_\rho(g)(x)\] 
has all degree-1 coefficients 0 and can be used as a control variate. 
%We can further reduce variance by subtracting the mean of this term to obtain.
%\[g(x) -\frac{T_\rho(g)(x)}{\rho}-\left(1-\frac{1}{\rho}\right)E[g]\] 

If $g$ is well correlated with $f$, then subtracting the low-variance $T_\rho(g)(x)/\rho$ term from $g(x)$ results in a function with 0 degree-1 coefficients with higher-order terms that are still well correlated with $f$. Subtracting this from $f$ to give $f(x)-(g(x) -T_\rho(g)(x)/\rho)$ then gives a function with small higher-order terms where the degree-1 terms are unaffected. A correlated function $g$ can be learnt by using a feedforward network and minimizing the mean squared error with $f$. 

Another possible way to build the control variate is to use
\[g(x) -T_\rho(g)(x)-T_{1-\rho}(g)(x).\] 

\subsection{Building the Estimator}
We built our final estimator used in the experiments upon MuProp. To do so we construct a function that includes the muProp terms, an input dependent baseline and our sample dependent baseline as follows.
\[ t(x) = f(x) - b(x) - f(\mu) - \alpha f'(x)(x-\mu)- \beta\left(g(x) - \frac{T_\rho(g)(x)}{\rho}\right) \]
The gradient is $\E[t(x)\nabla_\theta \log p(x)]$ and we use the single sample estimate of the expectation where $T_\rho(g)(x)$ is estimated using a single $\rho$-correlated sample.

\begin{table}[t]
\centering
\begin{tabular}{lccc}  
%\toprule
    \multicolumn{4}{c}{\bf{MNIST Generative Modeling}} \\
%\cmidrule(r)%{1-2}
    \midrule
    Model           & NVIL      & MuProp     & MuProp+Baseline \\
\midrule
    200-784         & -112.8    & -111.9    & -111.8          \\
    200-200-784     & -100.55         & -99.14    & -98.97             \\
    200-200-200-784 & -96.55    & -96.1     & -95.77             \\
\bottomrule
\end{tabular}
    \vspace*{5pt}
    \caption{Training ELBO for the MNIST dataset} \label{tab:mnist}
%\end{table}

%\begin{table}
\centering
\begin{tabular}{lccc}  
%\toprule
    \multicolumn{4}{c}{\bf{Omniglot Generative Modeling}} \\
%\cmidrule(r)%{1-2}
    \midrule
    Model           & NVIL      & MuProp       & MuProp+Baseline \\
\midrule
    200-784         & -118.7    & -118.3       & -118.4 \\
    200-200-784     & -110.8    & -110.2       & -109.44 \\
    200-200-200-784 & -108.8    & -107.94      & -107.76 \\
\bottomrule
\end{tabular}
    \vspace*{5pt}
    \caption{Training ELBO for the Omniglot dataset} \label{tab:omniglot}
\end{table}

\section{Experiments}
To test the variance reduction properties of our proposed estimator, we compared the estimator against NVIL and the MuProp estimator with an input dependent baseline. We built our implementation upon the code base made available by \cite{rebar}\footnote{TensorFlow code available at: \url{https://github.com/alpz/fourier-REINFORCE}}. In our experiments we used sigmoid belief nets with one, two and three layers. Each stochastic layer in the network is 200 units wide. The input dependent baseline is a feedforward newtork with a single layer of 100 tanh units and the sample-dependent baseline has two 100 unit layers of either tanh or relu units. The datasets are the statically binarized MNIST digit and Omniglot character datasets. For the optimization we used SGD with a momentum of 0.9 and a minibatch size of 24. For 3 layer models on Omniglot we found SGD to be slow to converge regardless of the gradient estimator; for those we performed the optimization using Adam. The gradient variance is estimated using an exponential moving average. We use a constant $\rho$ value of 0.5. The experiments were run with learning rates in $\{0.0001, 0.0003, 0.0006, 0.0009, 0.002\}$ and with the best performing result chosen. We used a different learning rate for the baseline networks which was set to $0.1$ times the learning rate for the model.

\subsection{Generative Modeling with Sigmoid Belief Nets}
We train generative models using the autoencoding variational Bayes framework of \cite{vae}. The framework simplifies the training of sigmoid belief networks using an inference network to generate samples from an approximate variational posterior distribution. The inference network is parameterized as a feedforward network with a structure that is the reverse of the model. We optimize the single sample variational lower bound (ELBO) of the log-likelihood.
\[ \log p(y|\theta) \ge \E_{q(x|y,\theta)} [ \log p(y|x,\theta) + \log p(x|\theta) - \log q(y|x,\theta)] \]
Here $q$ is the approximate variational posterior and $y$ is a data sample.

\begin{figure}
    \centering
    \begin{subfigure}{0.4\textwidth}
        \includegraphics[width=\textwidth]{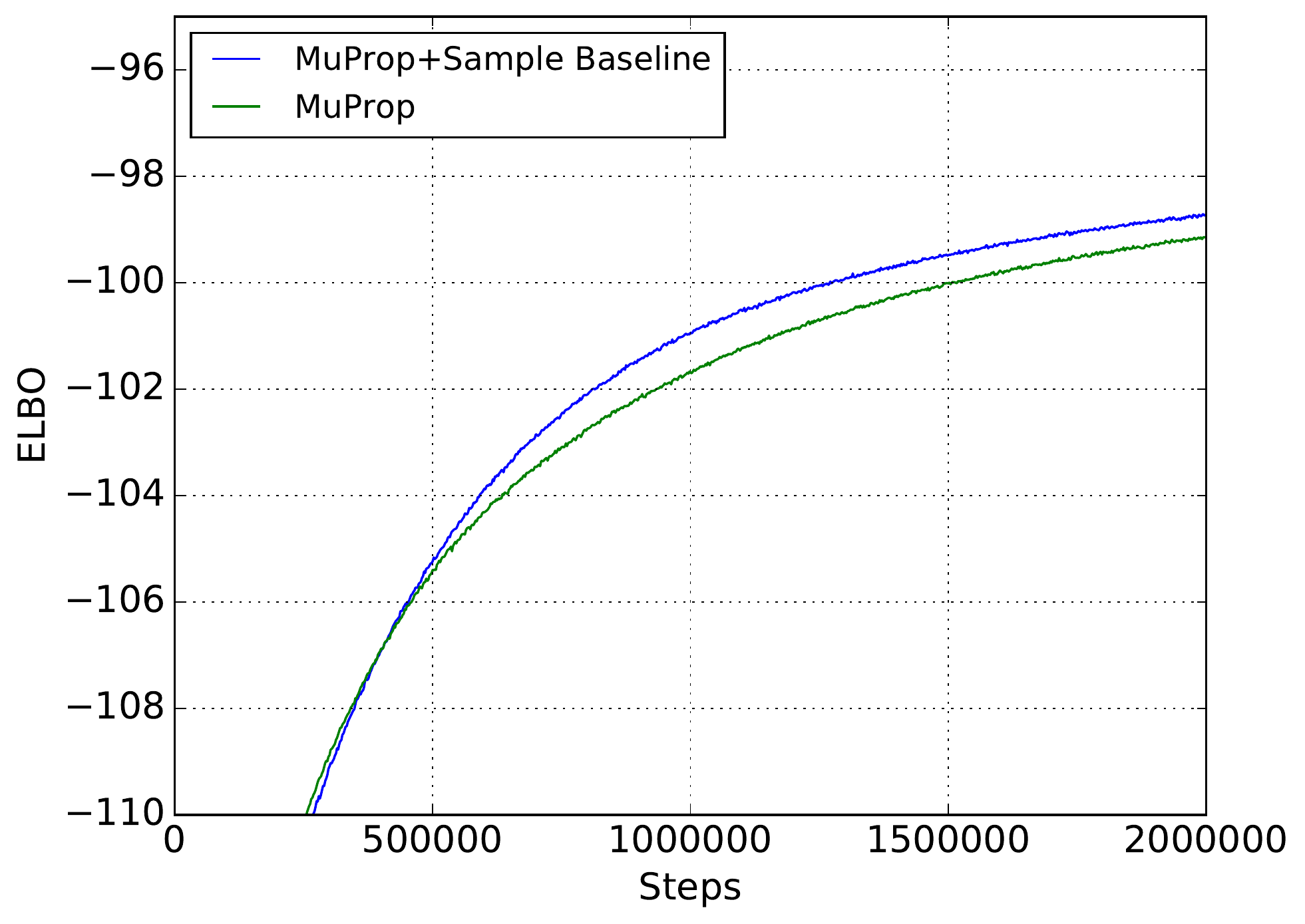}
    \end{subfigure}
    \begin{subfigure}{0.4\textwidth}
        \includegraphics[width=\textwidth]{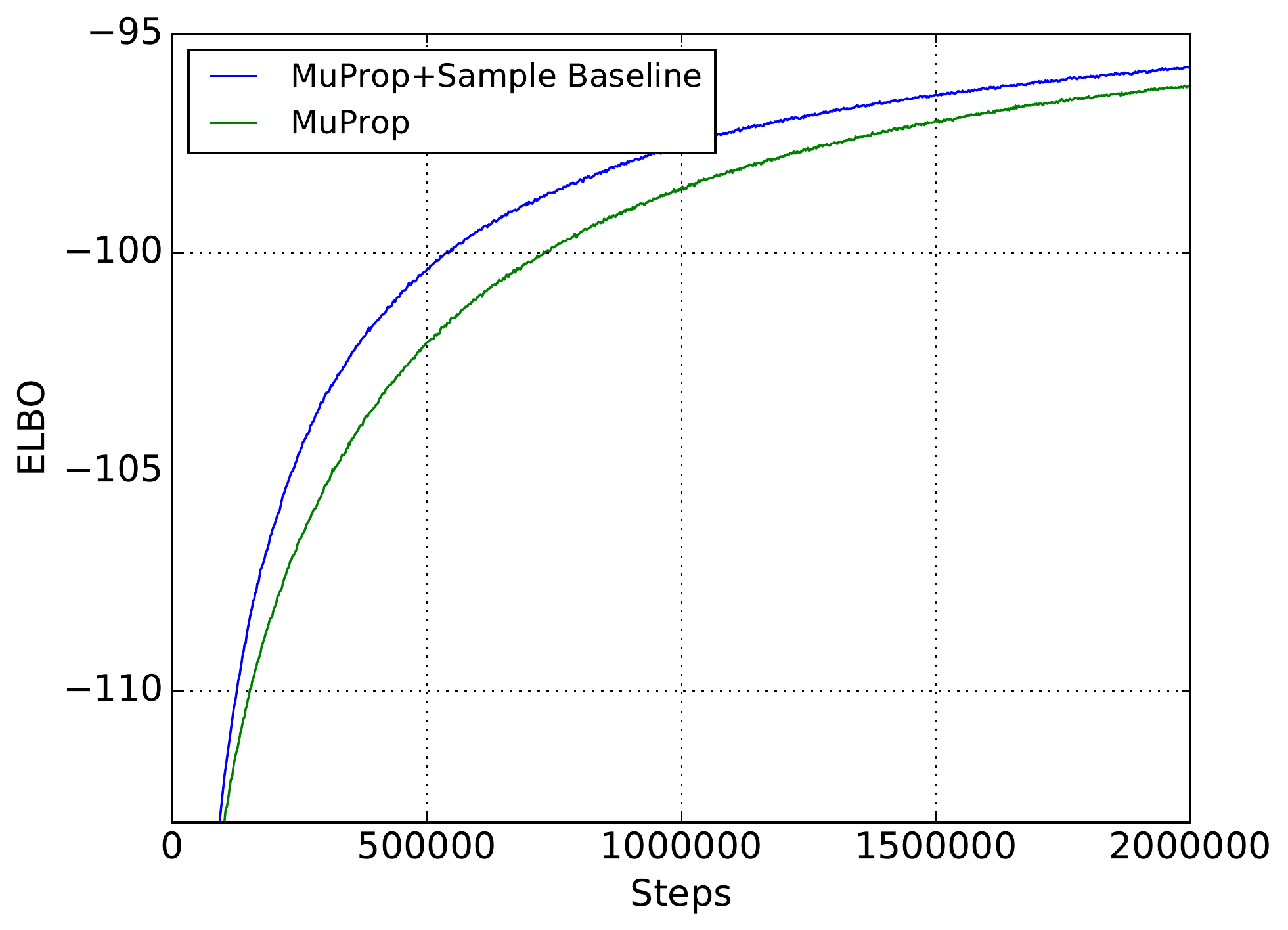}
    \end{subfigure}
    \caption{Training ELBO for 2 layer SBN (left) and 3 layer SBN (right) models on MNIST}
    \label{fig:elbo}
\end{figure}

\begin{figure}
    \centering
    \begin{subfigure}{0.4\textwidth}
        \includegraphics[width=\textwidth]{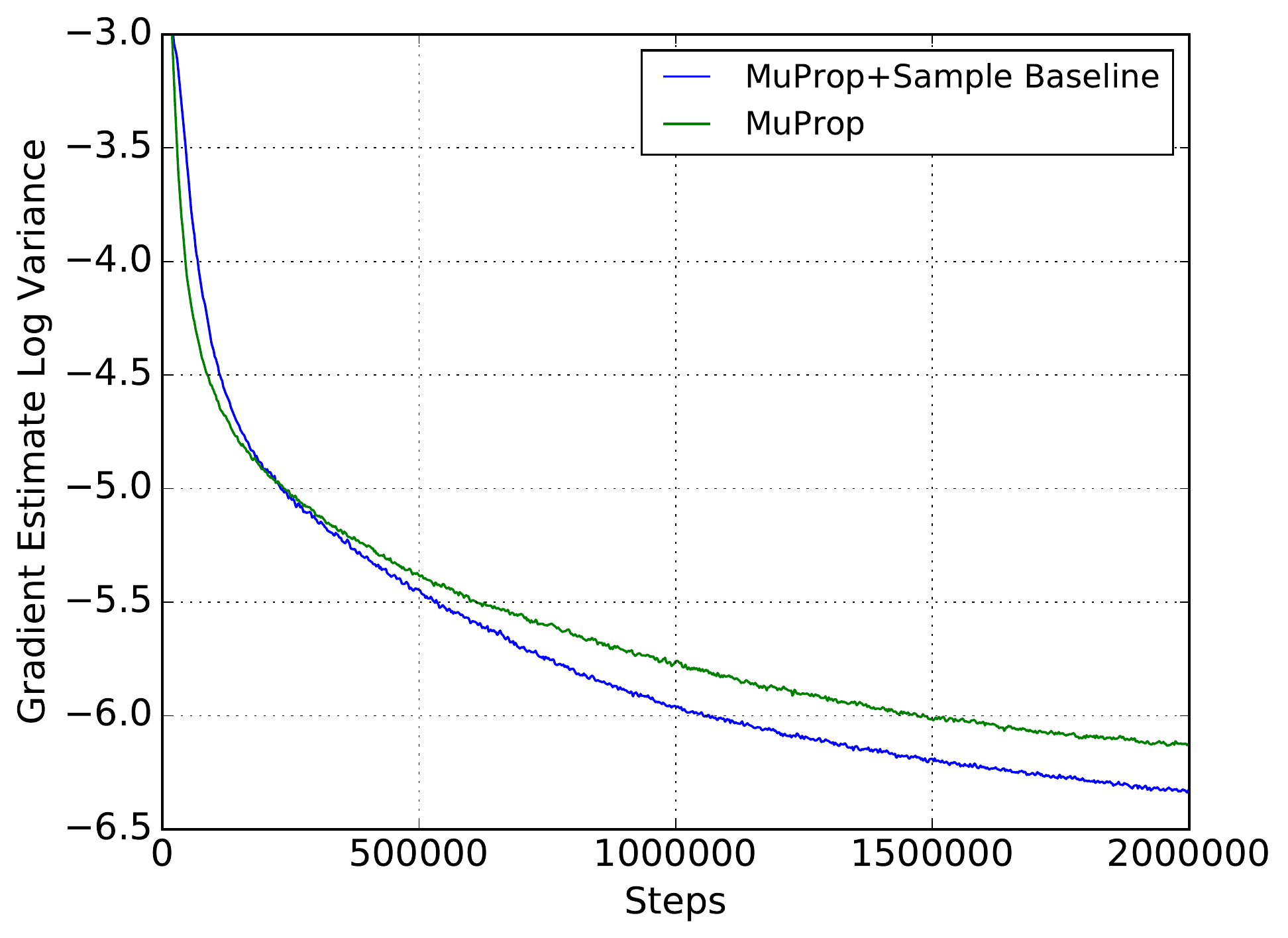}
    \end{subfigure}
    \begin{subfigure}{0.4\textwidth}
        \includegraphics[width=\textwidth]{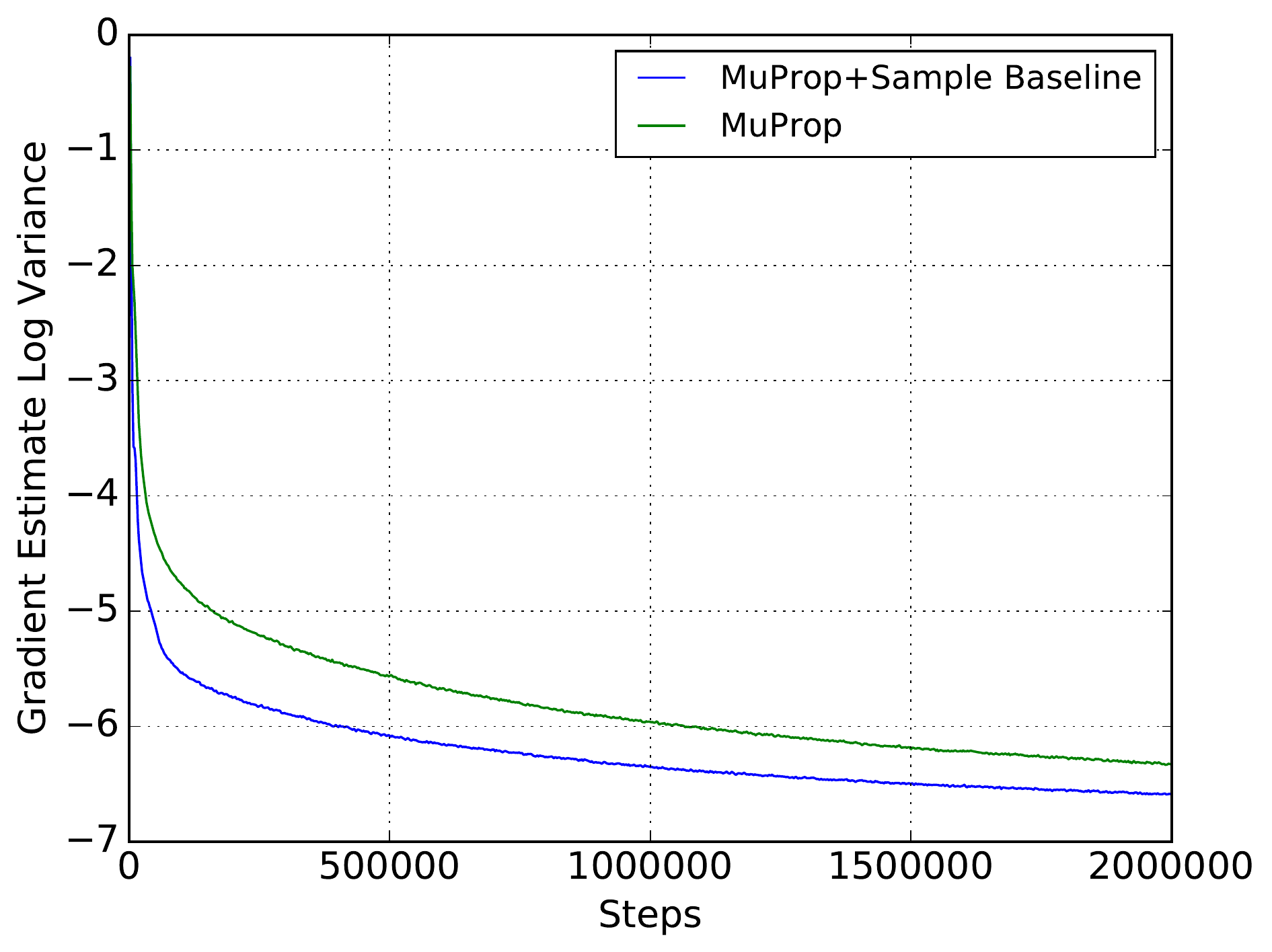}
    \end{subfigure}
    \caption{Log of gradient variance for 2 layer SBN (left) and 3 layer SBN (right) models on MNIST}
    \label{fig:grad}
\end{figure}

The final training ELBO results for MNIST and Omniglot after 2,000,000 steps are given in tables \ref{tab:mnist} and \ref{tab:omniglot}. The training ELBO for 2-layer and 3-layer models for MNIST is plotted in figure \ref{fig:elbo}. The estimated log of gradient variance is plotted in figure \ref{fig:grad}. As can be seen from the last figure we get a substantial improvement in gradient variance for the 2-layer model and the difference becomes significant early in the training when model depth is increased. This can also be seen from the ELBO plots where the 3-layer ELBO diverges from the MuProp ELBO much earlier than for the 2-layer model.

\subsubsection{Multiple Samples}
\begin{figure}[h]
    \centering
    \includegraphics[width=0.5\textwidth]{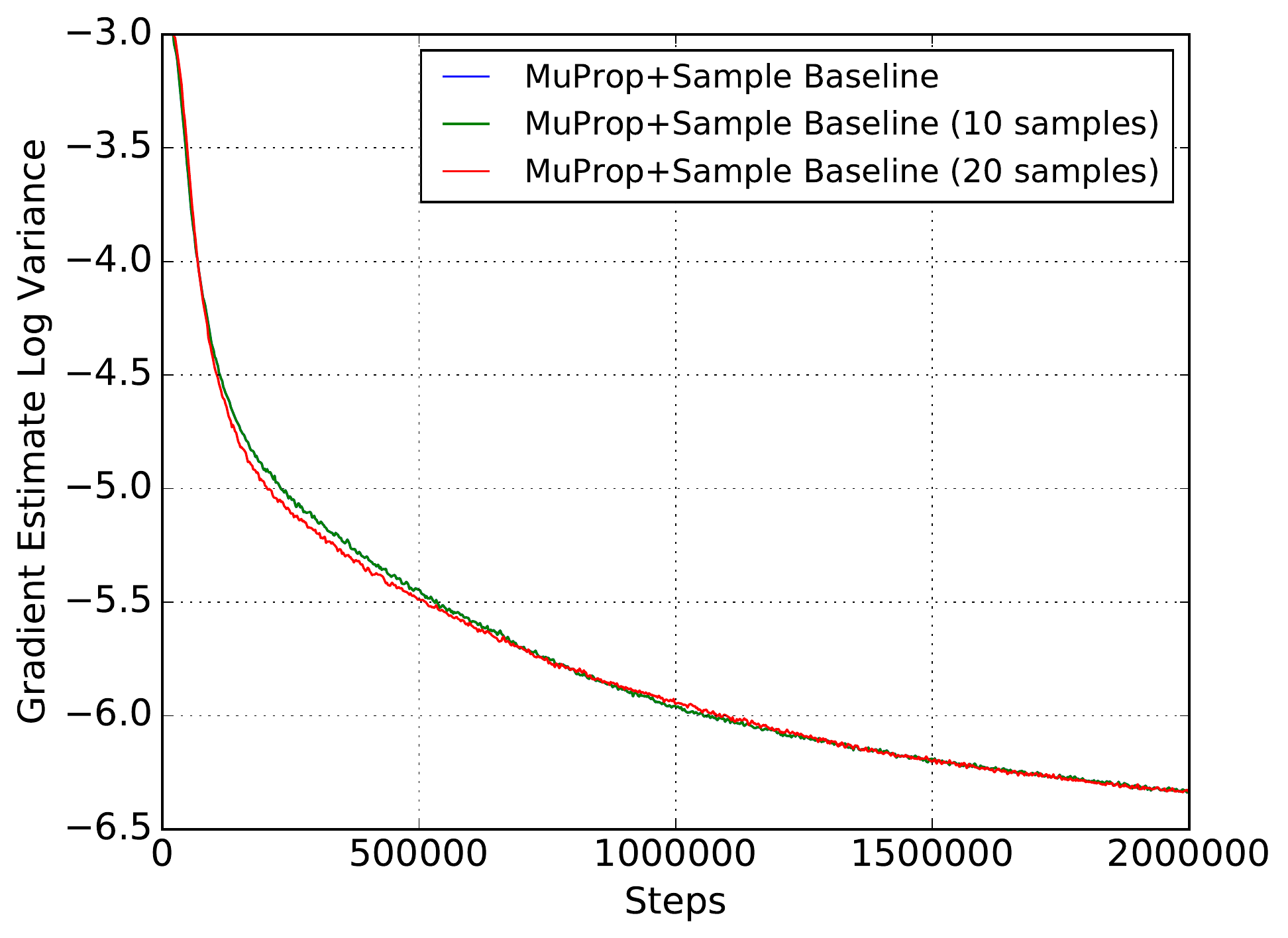}
    \caption{Log gradient variance for 2 layer SBNs with 1, 10 and 20 samples for computing $T_\rho(f)$ on MNIST.}
    \label{fig:samples}
\end{figure}

The results above were performed using a single sample to estimate $T_\rho(f)$. We also performed experiments using more samples to estimate $T_\rho(f)$. Notice that this still uses only a single sample from the model. This resulted in only very slight reduction in gradient variance in the initial phase of training. This can be seen in figure \ref{fig:samples} for the 2-layer model on MNIST. However since we did not optimize over hyperparameters for this, it still might be possible to improve these results for multiple samples to compute $T_\rho(f)$.

%\newpage

%[1] Alexander, J.A.\ \& Mozer, M.C.\ (1995) Template-based algorithms
%for connectionist rule extraction. In G.\ Tesauro, D.S.\ Touretzky and
%T.K.\ Leen (eds.), {\it Advances in Neural Information Processing
%  Systems 7}, pp.\ 609--616. Cambridge, MA: MIT Press.
%
%[2] Bower, J.M.\ \& Beeman, D.\ (1995) {\it The Book of GENESIS:
%  Exploring Realistic Neural Models with the GEneral NEural SImulation
%  System.}  New York: TELOS/Springer--Verlag.
%
%[3] Hasselmo, M.E., Schnell, E.\ \& Barkai, E.\ (1995) Dynamics of
%learning and recall at excitatory recurrent synapses and cholinergic
%modulation in rat hippocampal region CA3. {\it Journal of
%  Neuroscience} {\bf 15}(7):5249-5262.
\newpage
\normalsize
\appendix
\noindent{\Large\bfseries Appendix\par}

\section{Proof of Lemma \ref{lemma:1}}
We follow \cite[8.4]{odonnell} in the following proof.
  %  \[ \frac{\partial}{\partial p_i} E_{p_\theta (x)}[f(x)] = -\frac{2}{\sigma_i}\hat{f}(\{i\}), \]
\begin{proof}
    We consider $f$ as a multilinear polynomial over $x= (x_1, \ldots, x_n)$. Let $f^{(p)}$ be the $p$-biased Fourier representation of the Boolean function $f$. Then by linearity of expectation 
    \[ \E[f^{(p)}(x_1, \ldots, x_n)] = f(\mu_1, \ldots, \mu_n) \] 
    We also have that
    \[ \frac{\partial}{\partial \mu_i} f(\mu) = D_{x_i}f(\mu). \] 
    Then 
    \begin{align*}
        D_{x_i}f(\mu) &= \E[D_{x_i}f^{(p)}(x_1, \ldots, x_n)] \\
                    &= \frac{1}{\sigma_i}\E[D_{\phi_i}f^{(p)}(x_1, \ldots, x_n)] \\
                    &= \frac{1}{\sigma_i}\hat{f}(\{i\}),
    \end{align*}
    where $\hat{f}(\{i\})$ is the Fourier coefficient in the $p$-biased representation. Given that $\mu_i = 2p_i -1$, and $\frac{d\mu_i}{d p_i} = 2$, the result follows by the chain rule.
\end{proof}

\end{document}